%% file: arxiv.tex
\documentclass{article}

\usepackage[preprint]{neurips_2021}

\usepackage[utf8]{inputenc} % allow utf-8 input
\usepackage[T1]{fontenc}    % use 8-bit T1 fonts
\usepackage{hyperref}       % hyperlinks
\usepackage{url}            % simple URL typesetting
\usepackage{booktabs}       % professional-quality tables
\usepackage{amsfonts}       % blackboard math symbols
\usepackage{nicefrac}       % compact symbols for 1/2, etc.
\usepackage{microtype}      % microtypography
\usepackage{xcolor}         % colors
\usepackage{amsmath}
\usepackage{graphicx} 
\usepackage{array}
\usepackage{multirow}
\usepackage{multicol}
\usepackage{caption}

\def\eg{\textit{e.g.}}

\title{Co-advise: Cross Inductive Bias Distillation}

\author{Sucheng Ren$^{1,5}$~~ Zhengqi Gao$^{2}$~~ Tianyu Hua$^{4,5}$ ~~ Zihui Xue$^{3}$\\  ~~\textbf{Yonglong Tian$^{2}$~~Shengfeng He$^{1}$ ~~Hang Zhao$^{4,5}$ }
\thanks{Corresponding authors: Hang Zhao (hangzhao@mail.tsinghua.edu.cn), Shengfeng He (hesfe@scut.edu.cn)}\\
\hspace{-5mm}$^1$South China University of Technology ~~~   $^2$MIT\\
\hspace{-5mm}$^3$University of Texas at Austin ~~$^4$Tsinghua University ~~$^5$Shanghai Qi Zhi Institute
}
\begin{document}

\maketitle

\begin{abstract}

Transformers recently are adapted from the community of natural language processing as a promising substitute of convolution-based neural networks for visual learning tasks. However, its supremacy degenerates given an insufficient amount of training data (e.g., ImageNet). To make it into practical utility, we propose a novel distillation-based method to train vision transformers. Unlike previous works, where merely heavy convolution-based teachers are provided, 
%we introduce lightweight teachers specialized in different areas (e.g., convolution and involution) to co-advise the student transformer. 
we introduce lightweight teachers with different architectural inductive biases (e.g., convolution and involution) to co-advise the student transformer. 
The key is that teachers with different inductive biases attain different knowledge despite that they are trained on the same dataset, and such different knowledge compounds and boosts the student's performance during distillation. 
% Equipped with our cross inductive bias distillation method, our vision transformer (termed as CivT) outperforms all previous vision transformers of the same architecture on ImageNet.
Equipped with this cross inductive bias distillation method, our vision transformers (termed as CivT) outperform all previous transformers of the same architecture on ImageNet.

\end{abstract}

\input{1.intro_v2}

\input{2.related}
\input{3,method}

\input{4.exp}

\section{Conclusion}\label{section:conclusion}
% mention co-advice maybe?

In this paper, we introduce a cross inductive bias transformer (CivT) by distilling from teacher networks with diverse inductive biases. Compared with distilling from convolution teacher, cross inductive bias teachers provide different perspectives of data and avoid that student is over biased toward single teacher.
% Advised by two different teachers, a transformer student avoids being too biased towards the teacher in previous convolution-only knowledge distillation pipelines.
In our experiments, we find that the teacher inductive biases play a more critical role than the teacher performance in knowledge distillation. Based on this observation, only lightweight teachers is required in our method. Furthermore, we delve into the student model's inductive biases, and the capability of imitating teachers and the transformer shows its superiority in these two aspects comparing with Mixer and ResNet. Finally, we evaluate the effectiveness of the extra conv and inv tokens. They bring much performance gain but limited computation costs.

{
\small
\bibliographystyle{plain}
\bibliography{reference}
}

%%%%%%%%%%%%%%%%%%%%%%%%%%%%%%%%%%%%%%%%%%%%%%%%%%%%%%%%%%%%

%%%%%%%%%%%%%%%%%%%%%%%%%%%%%%%%%%%%%%%%%%%%%%%%%%%%%%%%%%%%

%\appendix

%\section{Appendix}

\end{document}

%% file: 1.intro_v2.tex
\section{Introduction}
\begin{figure}[t]
    \centering
    \begin{tabular}{c}
        \includegraphics[width=0.95\textwidth,height=3.5cm]{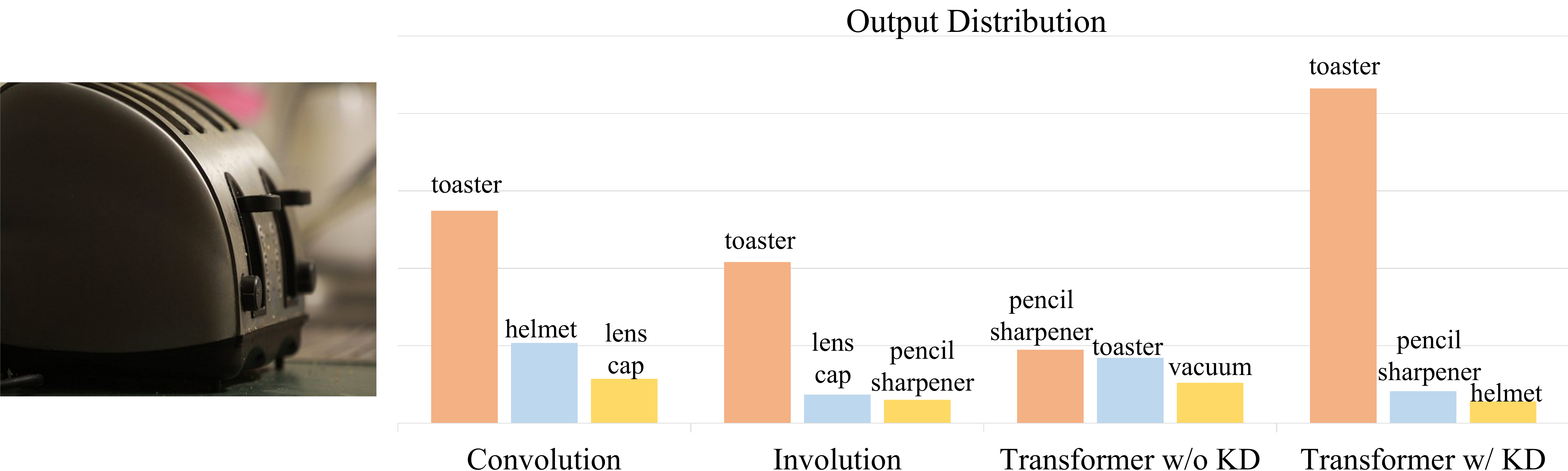}\\
        \includegraphics[width=0.95\textwidth,height=3.5cm]{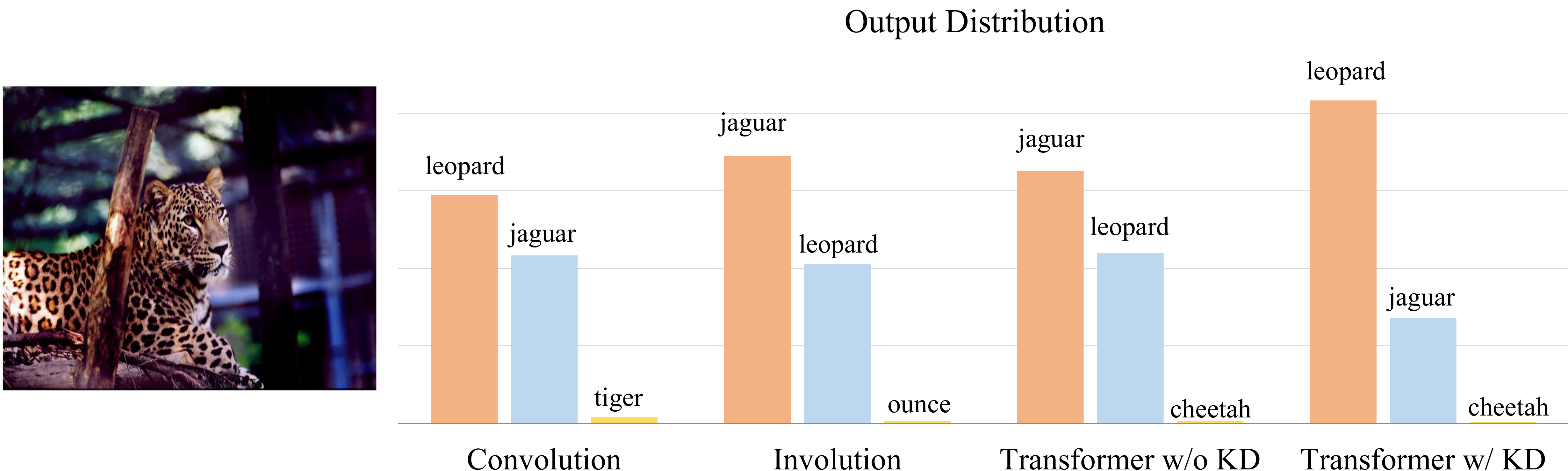}\\
        \includegraphics[width=0.95\textwidth,height=3.5cm]{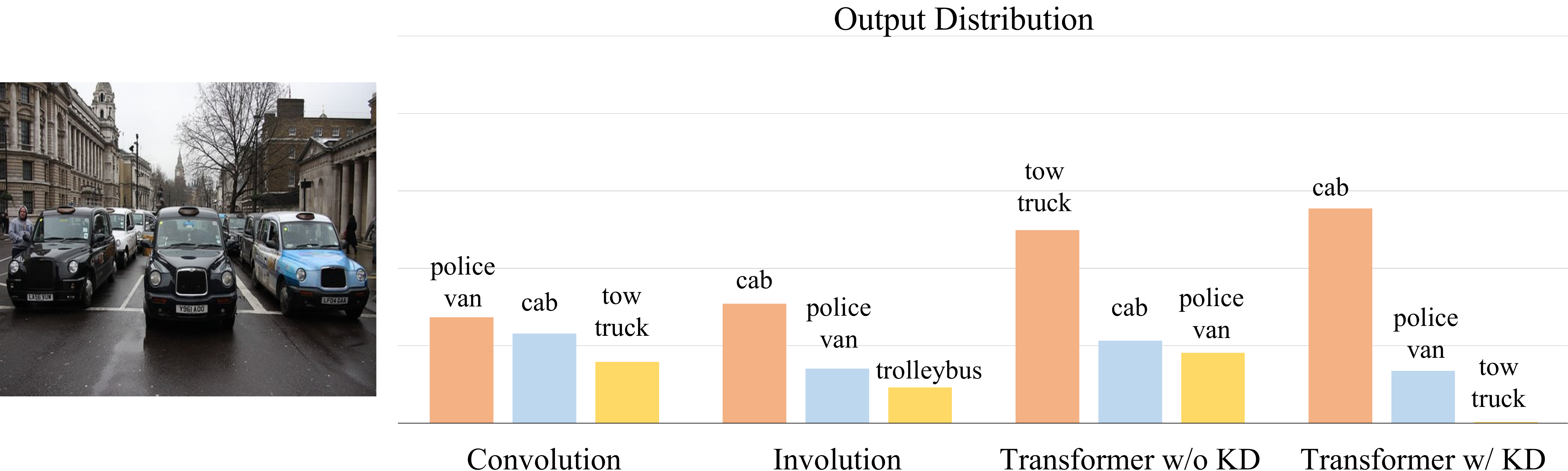}\\
    \end{tabular}\vspace{-3mm}
    \caption{Class probabilities predicted by a CNN, an INN, a transformer without distillation, and a transformer distilling from both CNN and INN. CNN and INN provide come up with consistent (the first row) or complementary (the second and third rows) conclusions to correct transformer's prediction.}
  \label{fig:cnn_inn_transf_bef_aft_distill}
\end{figure}

Though convolutional neural network (CNN) has revolutionized the field of computer vision, it possess certain limitations. The major one is that convolution operator processes signals in a spatial neighborhood, thereby making spatially-distant interactions difficult to be captured \cite{nonlocalnet,vt}. Motivated by this observation, vast research interests have been intrigued in replacing convolution layers with novel architectures \cite{nonlocalnet,involution,vit,detr,vt,deit}. Along this line of research, self-attention-based architectures (a.k.a.\ transformers \cite{attention_is_all_u_need}) form a notable mainstream \cite{vit,detr,vt} at present. For instance, ViT \cite{vit}, a pure transformer without convolutional layers, achieves state-of-the-art accuracy on ImageNet when pre-trained on large-scale datasets such as JFT-300M.

Nevertheless, when only trained with mid-sized datasets (e.g., ImageNet \cite{imagenet}), vision transformers typically achieves much lower accuracy than their convolutional counterparts~\cite{vit,deit}. The conjecture is that transformers has less inductive biases than CNNs (e.g., translation equivariance and locality) and thus suffers when the given amounts of training data is in sufficient~\cite{vit}. In this context, knowledge distillation technique \cite{hinton2015distilling,furlanello2018born} is applied by \cite{deit} to assist the training of transformers. As a result, transformers with such distillation~\cite{deit} (called DeiT) can achieve competitive results as CNNs on ImageNet, when the teacher CNN is powerful enough. Howerver, DeiT has its own limitations: the trained transformer is over-influenced by the inductive bias of the teacher CNN and mirrors CNN's classification error; DeiT requires the teacher CNN to be very large (e.g., RegNetY-16GF), which disturbingly brings about heavy computational overhead (e.g., training a RegNetY-16GF on ImageNet takes 6 days with 8 GPUs under the same training protocols in DeiT).

Recently, involution-based neural networks (INNs) emerges, and now we have three types of architectures (i.e., CNNs, transformers, INNs) for vision tasks. In this paper, we propose a cross inductive bias distillation method to train vision transformers, by leveraging both INNs and CNNs as teacher models. We believe that their inherent inductive biases (spatial-agnostic and channel-specific in convolution, spatial-specific and channel-agnostic in involution) could drive them to harvest different knowledge even though they are trained on the same dataset (see Figure \ref{fig:cnn_inn_transf_bef_aft_distill}). Therefore, teachers of different inductive biases could make different assumptions of the data and look at the data from different perspectives. In contrast, teachers with similar inductive biases but different performance (e.g., ResNet-18 and ResNet-50) have little difference in data description. Thanks to complementary inductive biases from different types of teachers, our method only requires two super lightweight teachers (a CNN and an INN), both of which are very easy to train. In the distillation stage, the knowledge from teachers compensates each other and significantly prompts the accuracy of the student transformer. The main contributions of this paper is four-fold:
\begin{itemize}
    \item Our cross inductive bias vision transformers (CivT) outperform all previous vision transformers of the same architecture on ImageNet and only requires super lightweight teachers whose total parameters are less than $20\%$ of those of DeiT-Ti and a half of those of DeiT-S respectively.
    % \item We observe that compared to the performance of a teacher model, its intrinsic inductive biases might even impact more on the student model.
    \item We observe that teacher model's intrinsic inductive bias matters much more than its accuracy.
    \item CNNs and INNs are inclined to learn texture and structure respectively, while a vision transformer, a more general architecture with fewer inductive biases, could inherit useful knowledge from both.
    \item When several teacher models with different inductive biases are provided, it is beneficial to choose a student model possessing few inductive biases.
\end{itemize}

The remainder of this paper is organized as follows. In Section \ref{section:related_work}, we briefly review some preliminary concepts. Next, we propose our cross inductive bias distillation method in Section \ref{section:proposed_method}. The efficacy of our method is thoroughly verified in Section \ref{section:experiment}. Finally, we conclude in Section \ref{section:conclusion}.

%% file: 2.related.tex
\section{Related Works}\label{section:related_work}

\paragraph{CNNs} Convolution operator was first proposed in \cite{LeNet} around thirty years ago. Its rejuvenation appears in the past decade, when deep CNNs (e.g., AlexNet \cite{AlexNet}, VGGNet \cite{VGG}, ResNet \cite{ResNet}, EfficientNet \cite{efficientnet}) led to an astonishing breakthrough in a great variety of tasks. The remarkable performance of CNNs origins from inherent characteristics (a.k.a.\ inductive biases) of the convolution operator such as translation equivariance \cite{vit} and spatial-agnostic \cite{involution}. On the other hand, its locality alternatively makes CNNs struggle to relate spatially-distant concepts, unless we deliberately increase the kernel size and/or model depth. % Please refer to \cite{vt} for a detailed discussions of CNNs' drawbacks.

\paragraph{Transformers} Transformers, which first prevailed in natural language processing \cite{attention_is_all_u_need}, has drawn attention in the computer vision community recently. The ViT proposed in \cite{vit} feeds $16\times 16$ image patches into a standard transformer, achieving comparable results as CNNs on JFT-300M \cite{vit}. However, its superiority is at the expense of excruciatingly long training time and tremendous amount of labeled data. Most importantly, when insufficient amount of data is given, ViT only achieves modest improvement of accuracy. Besides, DETR and VT were proposed in \cite{detr} and \cite{vt}, respectively. DETR \cite{detr} exploits bipartite matching loss and a transformer-based encoder-decoder structure in object detection task, while VT \cite{vt} represents images as semantic tokens and exploit transformers in image classification and semantic segmentation tasks. In addition to the aforementioned application perspective, it has been theoretically proven in \cite{cnn_attention_relation} that the self-attention mechanism used in transformers is at least as expressive as any convolution layer.

\paragraph{INNs} Involution operator was proposed in \cite{involution} lately. In a nutshell, convolution operator is spatial-agnostic and channel-specific, while an involution kernel is shared across channels and distinct in the spatial extent. In other words, involution attains precisely inverse inherent characteristics compared to convolution. As a result, it has the ability to relate long-range spatial relationship in an image. It is depicted in \cite{involution} that their involution-based RedNet consistently delivers enhanced performances compared with CNNs and transformers.

\paragraph{Knowledge Distillation} Knowledge distillation (KD) was first formulated in \cite{hinton2015distilling} as a strategy of model compression, in which a light-weight student is trained from a high-capacity teacher \cite{xue2021multimodal,tung2019similaritypreserving}. Specifically, authors in \cite{hinton2015distilling} achieve this goal by minimizing the KL divergence of student's and teacher's probabilistic predictions. Afterwards, KD unfolds usefulness in various tasks such as privileged learning \cite{unifying_distill_privileged_info, tung2019similaritypreserving}, cross-modal learning \cite{xue2021multimodal,cross_modal_ada_rgb_d}, adversarial learning \cite{adv_defense_distill,adv_attack_distill}, contrastive learning \cite{tian2020contrastive}, and incremental learning \cite{incre_learn_distill}. In relevance to our work, authors in \cite{deit} proposed to train transformers via a token-based KD strategy. By distilling from a large-scale and powerful CNN teacher, the resulting DeiT \cite{deit} can perform as well as CNNs, while the preceding ViT \cite{vit} cannot. Our method outperforms DeiT by distilling from two weak teachers with much fewer parameters, worse accuracy but different inductive bias.

%% file: 3,method.tex
\section{Vision Transformer with Cross Inductive Bias Distillation}\label{section:proposed_method}

\begin{figure}
  \centering
  \includegraphics[width=0.8\textwidth]{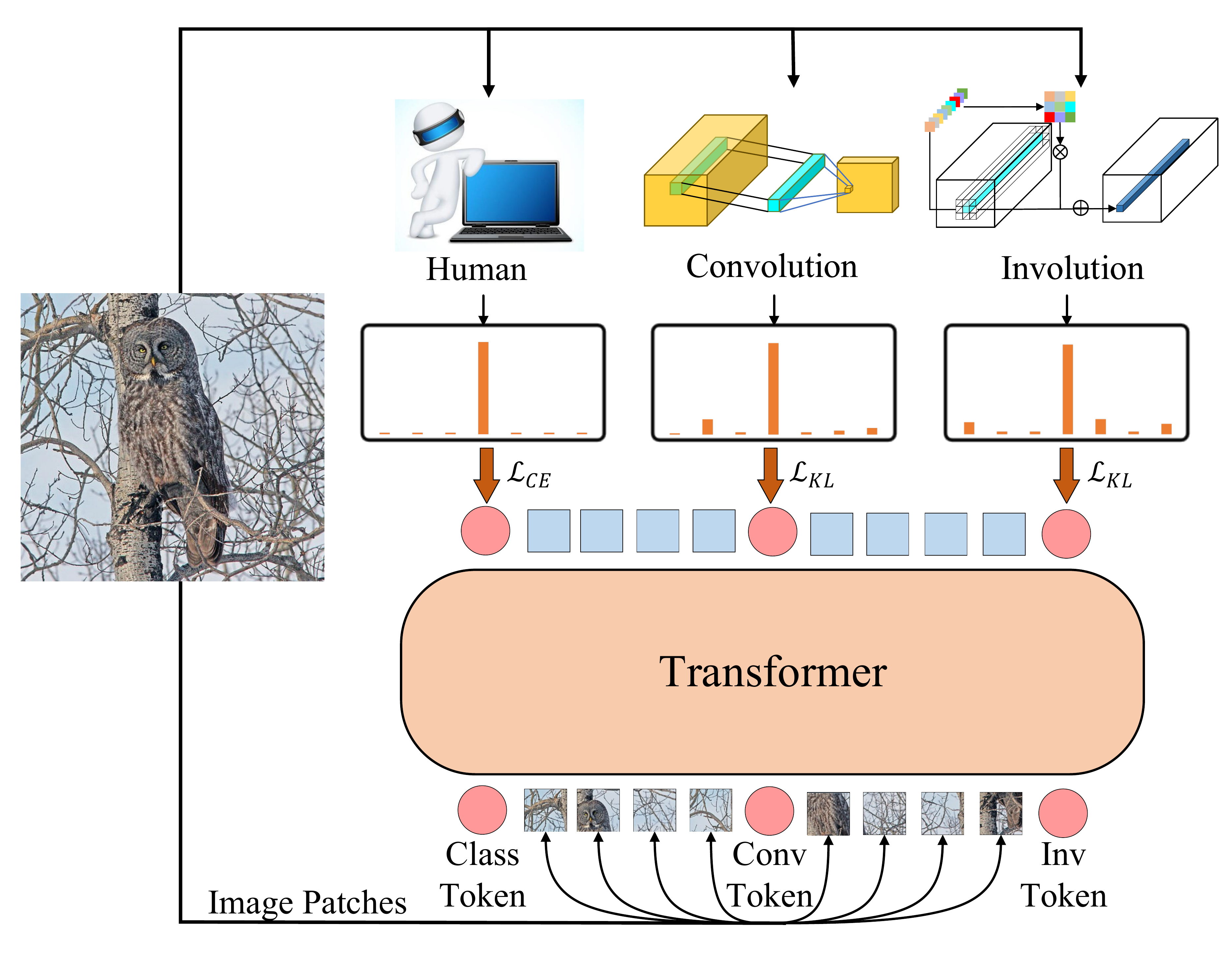}
  \caption{Schematic of our CivT. Given an image as input, human, convolution model and involution model will provide three similar (but slightly different) distributions to describe the image class. Our CivT model inherits the architecture of ViT but has two extra tokens (i.e., Conv token and Inv token) to learn from the convolution and involution teachers respectively.}
  \label{fig:schematic_civt}
\end{figure}

\subsection{Multi-Head Self-Attention Layer}
Suppose $N$ input feature vectors $\{\mathbf{x}_n \in \mathcal{R}^{d_{i}}\, |\, n=1,2,\cdots,N\}$ are stacked into the rows of a matrix $\mathbf{X}\in\mathcal{R}^{N\times d_i}$. A single-head self-attention layer first calculates the corresponding query, key and value matrices based on:
\begin{equation}
\begin{aligned}
         \mathbf{Q}=\mathbf{X}\mathbf{W}^Q\in\mathcal{R}^{N\times d_k}, 
         \quad\mathbf{K}=\mathbf{X}\mathbf{W}^K\in\mathcal{R}^{N\times d_k},
         \quad\mathbf{V}=\mathbf{X}\mathbf{W}^V\in\mathcal{R}^{N\times d_v},
\end{aligned}
\end{equation}
where $\{\mathbf{W}^Q\in\mathcal{R}^{d_i\times d_k},\,\mathbf{W}^K\in\mathcal{R}^{d_i\times d_k},\,\mathbf{W}^V\in\mathcal{R}^{d_i\times d_v}\}$ are parameters required to be learnt. The ultimate output of this single-head self-attention layer is given by:
\begin{equation}
    \text{Attention}(\mathbf{Q},\mathbf{K},\mathbf{V})=\sigma(\frac{\mathbf{Q}\mathbf{K}^T}{\sqrt{d_k}})\mathbf{V},
\end{equation}
where $\sigma(\cdot)$ represents the Softmax function, and is applied to each row of the matrix here. Built upon this, a multi-head self-attention layer is essentially an ensemble of $h$ independent self-attention layers. To be more specific, the $i$-th head generates a set of query, key and value matrices $\{\mathbf{Q}_i,\mathbf{K}_i,\mathbf{V}_i\}$ and gives an attention result:
\begin{equation}\label{eq:expression_head}
    \mathbf{head}_i=\text{Attention}(\mathbf{Q}_i,\mathbf{K}_i,\mathbf{V}_i)=\text{Attention}(\mathbf{X}\mathbf{W}^Q_i,\mathbf{X}\mathbf{W}^K_i,\mathbf{X}\mathbf{W}^V_i).
\end{equation}
Then the final output is given by fusing all query, key and value matrices $\{\mathbf{Q}_i,\mathbf{K}_i,\mathbf{V}_i\,|\,i=1,2,\cdots,h\}$ via concatenation and a linear transformation defined by $\mathbf{W}^O\in\mathcal{R}^{hd_v\times d_o}$:
\begin{equation}
    \text{MultiHead}(\mathbf{Q},\mathbf{K},\mathbf{V}) = \text{Concat}(\mathbf{head}_1,\cdots,\mathbf{head}_h)\mathbf{W}^O.
\end{equation}
Note that our notation in Eq (\ref{eq:expression_head}) is slightly different from the original expression shown in \cite{attention_is_all_u_need}, where, for instance, they denote $\mathbf{Q}_i=\mathbf{Q}\mathbf{W}^Q_i$ instead. However, we emphasize that these two notations are identical as performing two consecutive linear transformations can always be simplified as one.

\subsection{Sketch of Vision Transformer}

Following \cite{vit}, we first reshape the 2D image $\mathbf{x}\in\mathcal{R}^{H\times W\times C}$ into a sequence of image patches $\mathbf{x}_p\in\mathcal{R}^{M\times CP^2}$, where $H$ and $W$ are the height and width of the original image respectively, $C$ is the number of channels, and $P$ is the size of image patch. Once $P$ is determined, the number of patches $M$ is also determined as $M=HW/P^2$. Before being fed into the transformer, all image patches are first projected via a shared linear transformation. Next, these patch embeddings are added with a set of learnable position embeddings, so that all positional information is maintained. Finally, a class token is concatenated with these enhanced patch embeddings. The output from the class token is regarded as the image representation. In a downstream task such as image classification, we could stack a trainable multi-layer perceptron (MLP) on to the image representation and class probabilities are attained. 

The inner structure of a vision transformer is made up of several repeated identical blocks. Each block consists of an aforementioned multi-head self-attention layer and a feed-forward network, with Layernorm and residual connection add-on. Please refer to \cite{vit} for more details.

\subsection{Cross Inductive Bias Distillation}
DeiT~\cite{deit}, where the teacher model is a single convolution-based architecture, is limited by the knowledge of the teacher. A popular idea to go beyond the teacher performance is an ensembling of multiple teachers with different initializations~\cite{hinton2015distilling}. However, those teachers with the same architecture have same inductive biases, therefore offer similar perspectives of data. 

We propose to utilize teacher models possessing different inductive biases. Specifically, we choose a CNN and an INN as teachers. Therefore, we have to introduce two additional tokens corresponding to the convolutional and involutional teachers, respectively. The schematic of our CivT is demonstrated in Figure \ref{fig:schematic_civt}.

Our learning objective is expressed as a weighted sum of two Kullback-Leibler divergence losses ($\mathcal{L}_{KL}$) and a cross-entropy loss ($\mathcal{L}_{CE}$):
\begin{equation}\label{eq:loss}
    \mathcal{L} = \lambda_0 \mathcal{L}_{CE}(\sigma(\mathbf{z}_s),\mathbf{y})\,+\, \lambda_1\,\tau_1^2\,  \mathcal{L}_{KL}[\sigma(\frac{\mathbf{z}_s}{\tau_1}),\sigma(\frac{\mathbf{z}_{t_1}}{\tau_1})]\,+\, \lambda_2\,\tau_2^2\, \mathcal{L}_{KL}[\sigma(\frac{\mathbf{z}_s}{\tau_2}),\sigma(\frac{\mathbf{z}_{t_2}}{\tau_2})],
\end{equation}
where $0<\tau_1,\tau_2<\infty$ are hyper-parameters controlling the temperature of Softmax function $\sigma$ \cite{hinton2015distilling}. $\mathbf{z}_s$, $\mathbf{z}_{t_1}$ and $\mathbf{z}_{t_2}$ denote logits of the student transformer, CNN teacher and INN teacher, respectively. $0\leq \lambda_0,\lambda_1,\lambda_2\leq 1$ are weights balancing the importance of three loss terms.

%% file: 4.exp.tex
\section{Experimental Results}
\label{section:experiment}
In Section \ref{sec:imp_details}, we describe our implementation details, and next compare our CivT with various transformers, convolution- and involution-based neural networks on ImageNet-1k \cite{ImageNet_cvpr09} in Section \ref{sec:compare_with_sota}. In the rest of this section, experiments are conducted on ImageNet-100 \cite{wang2020hypersphere}. We analyze impacts of teacher performance and inductive biases to that of student transformer in Section \ref{sec:teacher_performance_inductive}. Then we explain the advantage of choosing a transformer as student over CNNs and INNs in Section \ref{sec:student}. To prove the effectiveness of our co-advising strategy, we compare the prediction accuracy of models trained by our cross inductive bias distillation and naive multi-teacher distillation in Section \ref{multiple_teacher}. Finally in Section \ref{sec:tokens}, we empirically prove the necessity of three tokens instead of one in our distillation method.

\subsection{Implementation Details}\label{sec:imp_details}
For comparison purpose, following DeiT \cite{deit}, we implement two variants of our model: (i) CivT-Ti has a hidden dimension of 192, 12 layers (each with three attention heads), and (ii) CivT-S has a hidden dimension of 384, 12 layers (each with six attention heads). We use the same data augmentation and regularization methods described in DeiT \cite{deit} (e.g., Auto-Augment, Rand-Augment, mixup). The weights of our transformers are randomly initialized by sampling from a truncated normal distribution. We use AdamW \cite{AdamW} as optimizer with learning rate equal to 0.001 and weight decay equal to 0.05. The whole training process includes 300 epochs. The first five epochs are for warm up, and in the remaining epochs the learning rate follows a cosine decay function. For hyper-parameters in distillation, we set $\lambda_0=\lambda_1=\lambda_2=1$ and $\tau_1=\tau_2=1$. During inference, we retrieve the value stored in the class token as the final output.

\subsection{Comparison among CNNs, INNs and Transformers}\label{sec:compare_with_sota}
\begin{table}[h]
\centering
\small
% \captionsetup{format=plain,labelsep=space,justification=raggedright,singlelinecheck=off,labelfont=bf}

\parbox{.4\textwidth}{\begin{tabular}{l|l|r|c}

\toprule
\multirow{2}{*}{Student} & \multicolumn{3}{c}{Teacher} \\
\cmidrule(lr){2-4}
         & Model    &Param      & Top-1 (\%)         \\
\midrule
DeiT     &    RegNetY-16GF& 84M& 82.9       \\
\midrule
\multirow{2}{*}{CivT-Ti}&   RegNetY-600M   &  6M   & 73.5\\
&RedNet-26&9M&73.0\\
\midrule
\multirow{2}{*}{CivT-S } &  RegNetY-4GF   &21M &  78.3 \\
&RedNet-50&16M&77.2\\
\bottomrule

\end{tabular}}
\hspace{\fill}
\parbox{.4\textwidth}{
\centering\caption{Comparison of teacher models used in DeiT \cite{deit} and CivT. DeiT uses a much larger and powerful convolution teacher, while CivT uses weak and small involution and convolution teachers.}
\label{tab:comparing_teachers}}
\end{table}

In this section, we compare accuracy of various convolution-, involution-, and transformer-based models on ImageNet-1k \cite{ImageNet_cvpr09}. 

\textbf{Teacher Model.} In Table \ref{tab:comparing_teachers}, we compare teacher models used in DeiT \cite{deit} and our CivT. Different from DeiT, which uses a powerful convolution teacher RegNetY-16GF \cite{regnet} with 84M parameters and top-1 accuracy of 82.9\%, we choose a convolution teacher and an involution teacher who possess similar model sizes as the student transformer. We emphasize that the overall parameters of teacher models used in our CivT are still much fewer than those in DeiT, and that such small teachers significantly speed up the whole training process.

\textbf{Results.} We report inference speed, top-1 accuracy of several models in Table \ref{tab:comparing_with_sota}. Compared with CNNs, when the model size is small (say around 6 million parameters), transformers do not reveal better performances. For instance, RegNet-600MF performs the best with top-1 accuracy equal to $76.3\%$, while DeiT-Ti, DeiT-Ti-KD, and our CivT-Ti achieve top-1 accuracy of 72.2\% ($-$4.1\%), 74.5\% ($-$1.8\%), and 74.9\% ($-$1.4\%), respectively. Namely, our CivT narrows the gap between the accuracy of CNNs and transformers in this context. When the model size grows, the accuracy of model grows much faster than that of other models, and our CivT-S outperforms all other models with about 20 million parameters. The performance of our CivT-S improves 2.6\% over RegNet-4GF and 2.9\% over RedNet-101.  

Compared with the recent transformer-based model ViT \cite{vit} (i.e., ViT-L /1 and ViT-B /16 in Table \ref{tab:comparing_with_sota}), our CivT-S requires about 4 times or 15 times fewer model parameters, while achieves about 4.1\% or 5.5\% more accurate predictions. Furthermore, our CivT-S also outperforms the latest work DeiT-KD, even though it has a more potent teacher. Specifically, the inference speed of DeiT-KD and our CivT are almost identical, while our CivT-Ti and CivT-S improve 0.4\% and 0.8\% over the corresponding DeiT-KD possessing similar sizes. To sum up, the extra convolution and involution tokens boosts the performance of student transformer almost without additional computation cost.

\begin{table*}[htp]
\small
\caption{Comparisons among CNNs, INNs and Transformers on ImageNet-1k \cite{ImageNet_cvpr09}. Throughput is measured on a single RTX3090 with batch size of 64.}
\centering
\small
%\resizebox{1.0\textwidth}{!}{
\begin{tabular}{l|l|r|r|r|c}
\toprule
%\multicolumn{1}{c}{\begin{tabular}[c]{@{}c@{}}throughput\\ (im/s)\end{tabular}} 
\multicolumn{2}{c|}{Model} & Param & Image Size & Throughput & Top-1 (\%)  \\
\midrule
     &ResNet-18~\cite{ResNet}        &  11.7M &  224$\times$224  &4595.8   &   69.8   \\
     &ResNet-50~\cite{ResNet}       &  25.6M   & 224$\times$224   &1349.4 &   76.2    \\
     %&ResNet-101~\cite{ResNet}       &  44.5M  &  224$\times$224 &799.4  &  77.4     \\
     &ResNet-152~\cite{ResNet}      &  60.2M  &  224$\times$224 &   555.9& 78.3   \\
     Convolution&RegNetY-600MF~\cite{regnet}   &  6.1M  & 224$\times$224&  1200.5   & 75.5    \\
     %Networks&RegNetY-800MF~\cite{regnet}     &  6.3M  &  224$\times$224 &  968.6   &  76.3    \\
     %&RegNetY-1.6GF~\cite{regnet}   &  11.2M &224$\times$224 &   695.7 &   78.0  \\
     %&RegNetY-3.2GF~\cite{regnet}   &  19.4M &  224$\times$224 &  694.4  &  79.0    \\
     Networks&RegNetY-4.0GF~\cite{regnet}     &  20.6M &224$\times$224  &  350.5&   79.4   \\
     &RegNetY-8.0GF~\cite{regnet}     &  39.2M &224$\times$224&  220.5&  79.9    \\

\midrule
%\multirow{5}{*}{Involution Networks}
   &RedNet-26~\cite{involution}&    9.2M      &224$\times$224  &    1820.9   &73.6   \\
    %&RedNet-38~\cite{involution}&     12.4M       & 224$\times$224    &    1369.9   &77.6   \\
    Involution&RedNet-50~\cite{involution}&    15.5M       & 224$\times$224   &   1066.8    &78.4      \\
    Networks&RedNet-101~\cite{involution}&     25.6M       &224$\times$224  &  657.4    &79.1     \\
    &RedNet-152~\cite{involution}&     34.0M      &  224$\times$224 &  459.3     &79.3    \\
\midrule
%\multirow{8}{*}{Transformer Networks}
    &ViT-B /16~\cite{vit}&      86M   & 384$\times$384  &     166.88      & 77.9    \\
    &ViT-L /16~\cite{vit}&   307M   &  384$\times$384 &   54.4     & 76.5      \\
    &DeiT-Ti~\cite{deit}&   5M         & 224$\times$224&      3082.9      &  72.2   \\
    Transformer&Deit-S~\cite{deit}&      22M      &224$\times$224 &    1562.0        &  79.8    \\
    Networks&DeiT-Ti-KD~\cite{deit}&   6M       & 224$\times$224&     3060.8   & 74.5    \\
    &DeiT-S-KD~\cite{deit}&         22M    &224$\times$224&       1546.1   & 81.2    \\
    &CivT-Ti (Ours)&      6M     & 224$\times$224 &        3053.0         &  74.9   \\
    &CivT-S (Ours)&       22M   &224$\times$224 &       1564.1         &   82.0  \\
\bottomrule
\end{tabular}%}
\label{tab:comparing_with_sota}
\end{table*}

\subsection{Teacher Performance and Inductive Biases}
\label{sec:teacher_performance_inductive}
% \begin{figure}
%   \centering
%   \includegraphics[width=1.0\textwidth]{figs/5.2.kd_differen_teacher.pdf}
%   \caption{Prediction accuracy of Transformer-Ti distilled from different teachers on ImageNet-100. Viewing horizontally reveals that the student's accuracy won't change much even though the teacher accuracy improves. Nevertheless, the vertical view demonstrates that teachers with same accuracy but belong to different kinds (e.g., CNNs or INNs) can yield students with different accuracy.}
%   \label{fig:different_teacher}
% \end{figure}

\begin{figure}[t]
    \centering
    \begin{tabular}{cc}
        \includegraphics[width=0.5\textwidth]{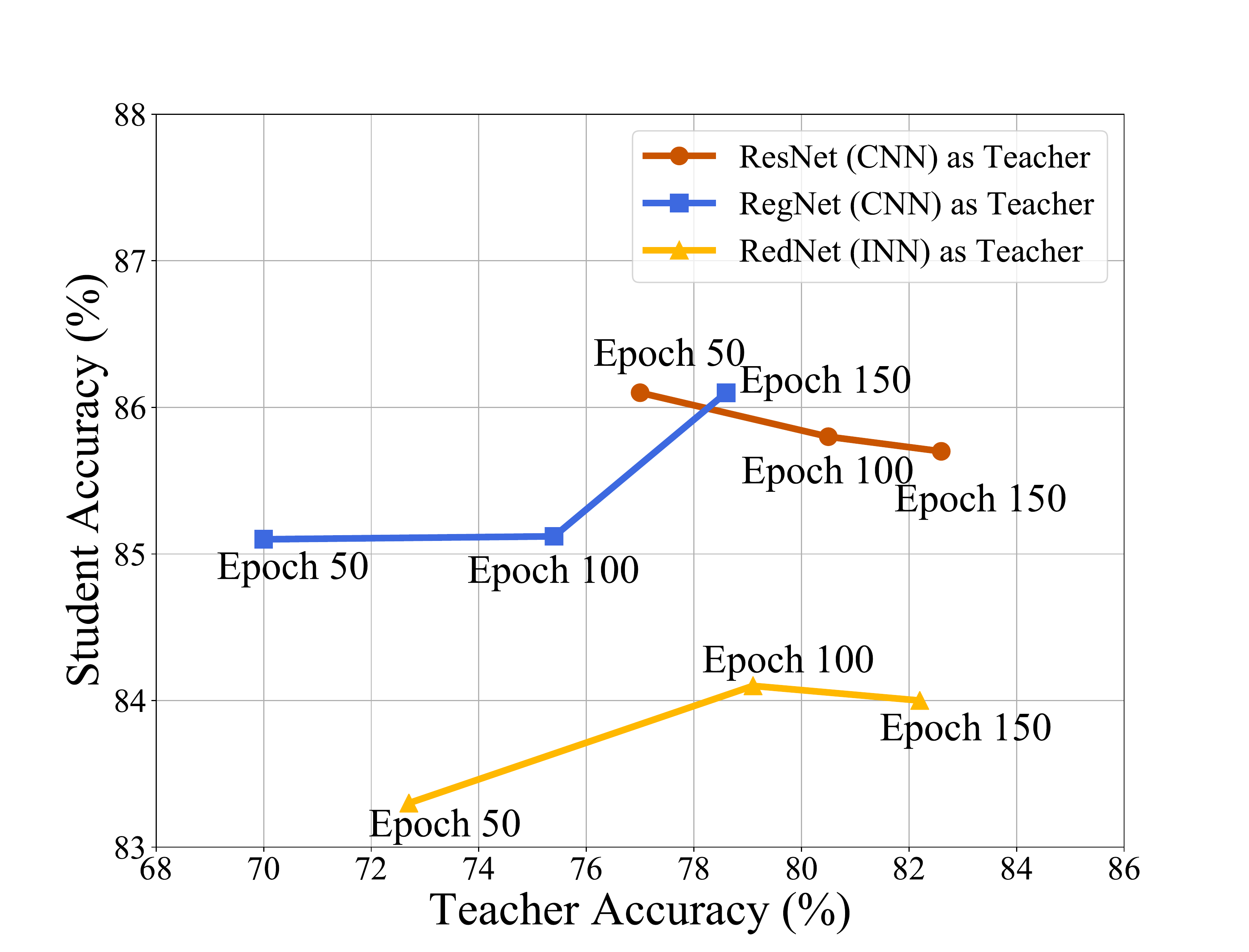}&
        \includegraphics[width=0.5\textwidth]{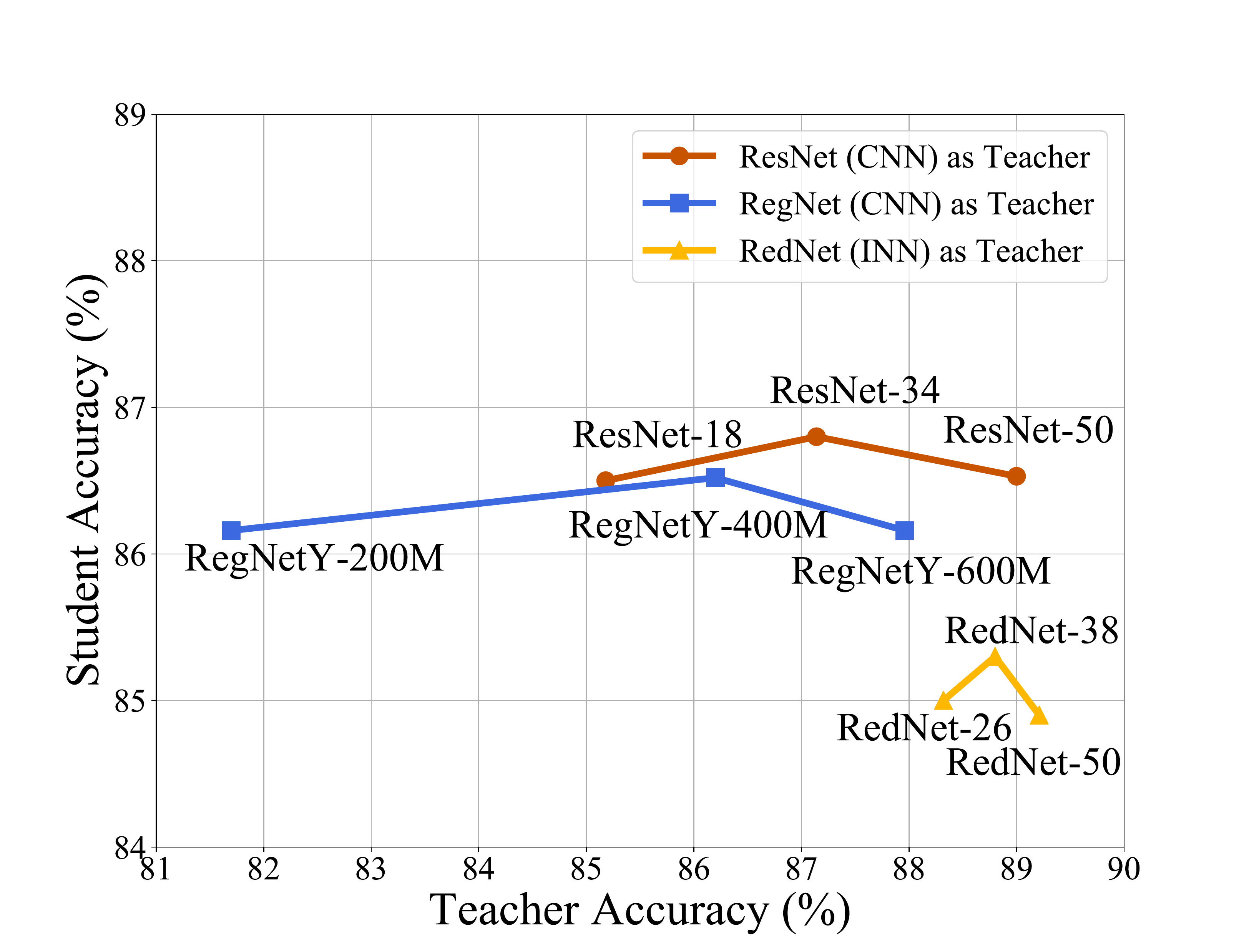}\\
        (a)&(b)\\
    \end{tabular}
    \caption{Prediction accuracy of Transformer-Ti distilled from different teachers on ImageNet-100. (a) We take ResNet-18, RegNetY-600M and RedNet-26 as teacher, the performance gap of teachers comes from different training epochs, but the students keep unchanged from horizontal view. (b) Viewing horizontally reveals that the student's accuracy won't change much even though the teacher accuracy improves. Nevertheless, the vertical view demonstrates that teachers with same accuracy but belong to different kinds (e.g., CNNs or INNs) can yield students with different accuracy.}
    \label{fig:different_teacher}
\end{figure}

This section delves into the impacts of teacher's performance and inductive biases when distilling to a student transformer. For illustration purpose, we conduct an experiment on student's accuracy when it distills from different kinds of teachers. We take three kinds of teachers into consideration: convolution-based ResNet and RegNet, and involution-based RedNet. We choose CivT-Ti as student. During distillation, only one teacher is provided, and thus one of the three tokens in CivT-Ti will be useless. From now on, this degenerated CivT-Ti will be referred to as Transformer-Ti. The results are reported in Figure \ref{fig:different_teacher}.

As shown in Figure \ref{fig:different_teacher}, if the teacher models share similar architecture (i.e., viewing horizontally in both (a) and (b)), the student model retains similar performance even though the teacher performances are boosted. For instance, in Figure \ref{fig:different_teacher}(a), different training epochs lead to the performance gap of teacher model. Training extra 100 epochs help the RegNet-200M teacher improve 9\%, but the performance of the student transformer keeps hardly changed. Similar observation can be generalized to ResNet-18 and RedNet-26 teachers. In Figure \ref{fig:different_teacher}(b) although the performances increase 6.5\% from RegNet-200M to RegNet-600M, the performance of students remains still. This observation implies that the accuracy of the teacher model is not the most important factor determining the student's performance in this context. Namely, we are approaching saturation: when the accuracy of teacher model is sufficiently large, improving teacher accuracy won't result in the improvement of student model. 

Alternatively, the vertical view of Figure \ref{fig:different_teacher} implies that we could resort to a teacher of a different type. For instance, when a teacher has similar performance but belongs to different kinds (\eg, ResNet-18 and RedNet-26 with training 150 epochs in Figure \ref{fig:different_teacher}(a), ResNet-50 and RedNet-50 in Figure \ref{fig:different_teacher}(b)), the distilled student could possess relatively different performances. Our hypothesis is that different kinds of teachers have different inductive biases. Even trained on the same dataset, they tend to harvest different knowledge. During distillation, some knowledge might be easier to be understood and inherited by the student model, while others do not. Furthermore, in terms of the student performance, the inherent knowledge of the teacher model seems to weigh more than its accuracy. \vspace{-3mm}
 
\subsection{Student Performance and Inductive Biases}
\label{sec:student}

\begin{table}[h]

\small
\centering
\caption{Performances of different students distilled from involution and convolution teachers. When both involution- and convolution-based teachers are provided, Transformer-Ti becomes CivT-Ti.}

\begin{tabular}{l|c|c|c}
\toprule

\multirow{2}{*}{Student} & \multicolumn{2}{c|}{Teacher}&\multirow{2}{*}{Top-1} \\
\cmidrule(lr){2-3}
         & ResNet-18 & RegNet-26 &   (\%)      \\
\midrule
ResNet-10     &    &  &     81.5   \\
ResNet-10     & \checkmark   & &    83.0    \\
ResNet-10     &    & \checkmark&     82.6   \\
ResNet-10     &  \checkmark  & \checkmark&   83.4     \\
\midrule
Mixer-Ti&       &     &80.5 \\
Mixer-Ti&    \checkmark   &     &81.6 \\
Mixer-Ti&       &  \checkmark   & 80.9\\
Mixer-Ti&   \checkmark    &  \checkmark   & 82.3\\

\midrule
Transformer-Ti&       &     & 81.8\\
Transformer-Ti&    \checkmark   &     &86.5 \\
Transformer-Ti&       &  \checkmark   &85.0 \\
Transformer-Ti (Ours)&   \checkmark    &  \checkmark   &88.0 \\

\bottomrule
\end{tabular}
\label{tab:comparing_student}
\end{table}

When distilling cross inductive knowledge to a student, the student needs to have few inductive biases to avoid overly inclining to a certain teacher. Moreover, the student model needs to have enough capability and model capacity to learn from its teachers. Based on these two considerations, we choose ResNet-10, Transformer-Ti, and Mixer-Ti \cite{mixer} as students and ResNet-18, RedNet-26 as teachers. ResNet-10 has stronger inductive biases than Transformer-Ti, and such inductive biases are similar to those of ResNet-18 and conflicts with those of RedNet-26. The results are reported in Table \ref{tab:comparing_student}.

Our experiment results demonstrate that ResNet-10 distilling from two teachers attains a similar performance to that distilling from a single convolution-based ResNet-18. In contrast, Transformer-Ti can learn from both teachers and achieve higher performance (88\%) than distilling from a single teacher. We believe the intrinsic reason is that a transformer possesses few inductive biases and the attention layer could not only perform convolution~\cite{Cordonnier2020On}, but also has close relationship to involution~\cite{involution}. 

This rises a natural question: An MLP possesses the fewest inductive biases, how about choosing it as student?  To this end, we include the recent Mixer model \cite{mixer}, a pure multi-layer perceptron (MLP) structure, into comparison. For fairness of comparison, the Mixer-Ti used in our paper has 12 layers, and the hidden dimension is 192. As shown in Table \ref{tab:comparing_student}, it indicates that without any distillation, Mixer-Ti and Transformer-Ti have similar performances. However, after distilling knowledge from teachers, Transformer-Ti gains more improvement than Mixer. This demonstrates the effectiveness of choosing transformer as a student.

The reason why Mixer-Ti doesn't gain as much as a Transformer through distillation will be clear if we compute the KL divergence between student's and teacher's outputs. As shown in Table \ref{tab:KL}, all values of KL divergence in Mixer-Ti are much larger than the others. It implies that Mixer-Ti doesn't have the ability to learn from the teacher when its model size is constrained to the same as its Transformer counterpart. On the contrary, compared with other students, CivT-Ti are more similar to teachers. Not surprisingly, the convolution token and involution token are more inclined to convolution and involution teacher, respectively, because our loss function in Eq (\ref{eq:loss}) advocates them to mimic their corresponding teachers.

\begin{table}[h]
\small
\centering
\caption{The output KL divergence. A smaller value indicates a larger similarity.}
\begin{tabular}{l|c|c|c}
\toprule
Student & ResNet-18& RedNet-26&Top-1 (\%) \\
\midrule
ResNet-10     &  0.261  & 0.274 &   83.4     \\
\midrule
Mixer-Ti&  0.358     &  0.313   &82.3 \\
\midrule
%CivT-Ti Class token&  0.329    &   0.282  & 88.0\\
CivT-Ti Conv token&    0.255   &  0.290   &87.1\\
CivT-Ti Inv token&   0.254    &   0.154  &87.7 \\

\bottomrule
\end{tabular}
\label{tab:KL}
\end{table}
\vspace{-2mm}

\subsection{Cross Inductive Bias Distillation and Naive Multi-Teacher Distillation}
\label{multiple_teacher}
In this section, we verify the effectiveness of our cross inductive bias distillation by comparing it with naive multi-teacher distillation. We implement three teachers: (i) ResNet-18, ResNet-50 are both convolution-based models. They have similar inductive biases, but different performances due to different model sizes. (ii) RedNet-26 is an involution-based model but with similar performance as ResNet-50. The results are illustrated in Table \ref{tab:comparing_teacher_inductive_bias}.

When Transformer-Ti distills from a single teacher, its performance gain is significant regardless the type of teacher. Specifically, after distilling from the convolution-based ResNet-18, Transformer-Ti can achieve about 86.5\% top-1 accuracy on ImageNet-100, while after distilling from the involution-based RedNet-26, its performance gain is relatively moderate: achieving 85.0\% top-1 accuracy. 

When one more teacher is further allowed in distillation, interesting phenomenon occurs. If both teachers are convolution based (a.k.a.\ teacher ensembling \cite{teacher_ensemble}), the further performance improvement is limited (\eg~from 86.5\% to 87.0\% or 87.2\%). In contrast, if we choose the additional teacher as the involution-based RedNet-26, the performance of Transformer-Ti rises to 88.0\%. This justifies the effectiveness of providing two different types of teachers.

\begin{table}[h]
\small
\centering
\caption{Performances of various models on ImageNet-100. A check mark $\checkmark$ represents a teacher of the specified type is presented. $\checkmark\checkmark$ indicates two architectural identical teachers with different initialization.}
\begin{tabular}{l|c|c|c|c}

\toprule
\multirow{2}{*}{Student} & \multicolumn{3}{c|}{Teacher}&\multirow{2}{*}{Top-1} \\
\cmidrule(lr){2-4}
         & ResNet-18 & ResNet-50& RedNet-26 & (\%)       \\
\midrule
ResNet-18&       &     && 85.1\\
ResNet-50&       &     && 89.0\\
RedNet-26&       &     && 89.2\\
Transformer-Ti&       &     && 81.8\\
\midrule
Transformer-Ti&    \checkmark   &  &   & 86.5\\
Transformer-Ti&      & \checkmark   &   &86.6 \\
Transformer-Ti&      &    &  \checkmark & 85.0\\
Transformer-Ti&    \checkmark  \checkmark  &  &   &87.2 \\
Transformer-Ti&      \checkmark  &  \checkmark  & &87.0 \\
Transformer-Ti (Ours)&   \checkmark    &    & \checkmark&88.0 \\

\bottomrule
\end{tabular}
\label{tab:comparing_teacher_inductive_bias}
\end{table}

\subsection{Effectiveness of Multiple Distillation Tokens}
\label{sec:tokens}
In conventional knowledge distillation \cite{hinton2015distilling}, one output token is used to fit the true label and teacher's logits simultaneously. However, such two objectives are sometimes in conflict~\cite{cho2019efficacy}. As shown in Eq (\ref{eq:loss}), we use different tokens to capture different knowledge provided by different teachers. Specifically, class , convolution and involution token learn from the true label, convolution teacher, and involution teacher, respectively. To evaluate the effectiveness of three tokens, we compare the accuracy of the learned Transformer with that trained via only one or two tokens. The results are reported in Table \ref{tab:tokens}. When the number of tokens is one, distilling from two teachers with different inductive biases can bring considerable improvements, while only distilling from one teacher induces almost no positive result. With the same teachers, merely by increasing from one token to three, our method achieves a 4.5\% accuracy improvement.

\begin{table}[h]
\small
\centering
\caption{Performances of various models on ImageNet-100. A check mark $\checkmark$ represents a teacher of the specified type is presented.}
\begin{tabular}{l|c|c|c|c}

\toprule
\multicolumn{2}{c|}{Student}& \multicolumn{2}{c|}{Teacher}&\multirow{2}{*}{Top-1} \\
\cmidrule(lr){1-2}\cmidrule(lr){3-4}
Model    & Token (s)   & ResNet-18 & RedNet-26& (\%)\\
\midrule
Transformer-Ti&   1   &   && 81.8\\
Transformer-Ti&   1  &\checkmark   &   & 81.9\\
Transformer-Ti&    1 &   &  \checkmark  & 80.7\\
Transformer-Ti& 1 & \checkmark   &  \checkmark &83.5\\
% \midrule
% Transformer-Ti&  2&  \checkmark   &   &86.5 \\
% Transformer-Ti&   2&    &\checkmark &85.0 \\
% Transformer-Ti&   2&   \checkmark  &  \checkmark &83.3\\
\midrule
Transformer-Ti (Ours)& 3   & \checkmark   & \checkmark&88.0 \\

\bottomrule
\end{tabular}
\label{tab:tokens}
\end{table}